%% file: main.tex
\documentclass[sigconf]{acmart}
\usepackage{multirow}
\usepackage{subcaption}

\copyrightyear{2026}
\acmYear{2026}
\setcopyright{cc}
\setcctype{by}
\acmConference[CIKM '26]{Proceedings of the 35th ACM International Conference on Information and Knowledge Management}{November 7--11, 2026}{Rome, Italy.}
\acmBooktitle{Proceedings of the 35th ACM International Conference on Information and Knowledge Management (CIKM '26), November 7--11, 2026, Rome, Italy}
\acmISBN{979-8-4007-2539-5/2026/11}
\acmDOI{10.1145/3799682.3840163}

\begin{document}

\title{LFM4Ads: Large Foundation Model for Ads Recommendation}
\input{0.authors}
\begin{abstract}

Recent work leverages large-scale pretrained foundation models (FMs) to transfer generalizable representations to downstream advertising tasks.
Most existing approaches transfer user representations (URs) and item representations (IRs), while underutilizing the user-item cross representations (CRs) generated by deeper interaction layers.
Although a recent work transfers target-aware embeddings (TAEs), it requires real-time upstream FM inference alongside the downstream, introducing additional engineering complexity and overhead.

To address these limitations, we propose Large Foundation Model for Ads Recommendation (LFM4Ads), an efficient \emph{pretrain-then-transfer} paradigm for cross-domain ad recommendation.
LFM4Ads jointly exploits URs, IRs, and CRs from the upstream FM.
It extracts sample-level CRs from pretrained interaction layers and aggregates them into user-level CRs using a time-aware exponential moving average, thereby decoupling upstream inference from downstream online serving.
To use CRs efficiently, we propose three downstream usages at different levels of granularity: feature-level gating to fuse CRs into downstream embeddings, module-level fine-tuning to yield downstream CRs, and model-level retrieval by matching CRs with IRs.

LFM4Ads has been deployed on Tencent's advertising platform since Q4 2024.
It has supported 10+ launches across major scenarios, yielding a platform-wide gross merchandise value (GMV) lift of 2.45\% and annual revenue gains in the hundreds of millions of dollars.
Offline experiments on KuaiRand-1K further demonstrate our effectiveness.
Our code is available at \textcolor[HTML]{0969DA}{\url{https://github.com/ZhuangQu/LFM4Ads}}.

\end{abstract}



\maketitle
\input{1.introduction}
\input{2.background}
\input{3.method}
\input{4.offline}
\input{5.online}
\input{6.conclusion}
\bibliographystyle{ACM-Reference-Format}
\balance
\bibliography{reference}

\end{document}

%% file: 0.authors.tex
\author{Jie Jiang}
\orcid{0000-0001-9658-5127}
\authornote{Equal contribution.}
\email{zeus@tencent.com}

\author{Shangyu Zhang}
\orcid{0000-0002-8614-6371}
\authornotemark[1]
\email{vitosyzhang@tencent.com}

\author{Shijie Quan}
\orcid{0009-0007-9154-4781}
\authornotemark[1]
\email{justinquan@tencent.com}
\affiliation{
\institution{Tencent}
\city{Shenzhen}
\country{China}
}

\author{Zhongren Wang}
\orcid{0009-0007-2178-3920}
\authornotemark[1]
\email{wcr23@mails.tsinghua.edu.cn}

\author{Tianqu Zhuang}
\orcid{0009-0006-8730-1878}
\authornotemark[1]
\email{zhuang-qu@qq.com}
\affiliation{
\institution{Tsinghua University}
\city{Shenzhen}
\country{China}
}

\author{Junwei Pan}
\orcid{0009-0003-2697-7012}
\authornote{Corresponding authors.}
\email{jonaspan@tencent.com}

\author{Bo Fu}
\orcid{0009-0004-0182-0999}
\email{leonardfu@tencent.com}

\author{Yilong Sun}
\orcid{0009-0007-2534-962X}
\email{elonsun@tencent.com}
\affiliation{
\institution{Tencent}
\city{Shenzhen}
\country{China}
}

\author{Jieying Lin}
\orcid{0009-0007-0939-114X}
\email{jieyinglin@tencent.com}

\author{Jushuo Chen}
\orcid{0000-0001-7546-2881}
\email{jushuochen@tencent.com}

\author{Xiaotian Li}
\orcid{0009-0000-7547-8525}
\email{dylanxtli@tencent.com}
\affiliation{
\institution{Tencent}
\city{Shenzhen}
\country{China}
}

\author{Wenjie Li}
\orcid{0000-0003-2717-9031}
\email{liwj20@mails.tsinghua.edu.cn}
\affiliation{
\institution{Tsinghua University}
\city{Shenzhen}
\country{China}
}

\author{Zhixiang Feng}
\orcid{0009-0005-1519-0227}
\email{lionelfeng@tencent.com}

\author{Xian Hu}
\orcid{0009-0002-4036-3968}
\email{sirlyhu@tencent.com}
\affiliation{
\institution{Tencent}
\city{Shenzhen}
\country{China}
}

\author{Huiting Deng}
\orcid{0009-0002-1649-6007}
\email{timmydeng@tencent.com}

\author{Hua Lu}
\orcid{0009-0009-2658-6613}
\email{hualu@tencent.com}
\affiliation{
\institution{Tencent}
\city{Shenzhen}
\country{China}
}

\author{Jinpeng Wang}
\orcid{0000-0002-4352-4897}
\email{wangjp26@gmail.com}
\affiliation{
\institution{Tsinghua University}
\city{Shenzhen}
\country{China}
}

\author{Boqi Dai}
\orcid{0009-0008-3796-2541}
\email{ericbqdai@tencent.com}

\author{Xiaoyu Chen}
\orcid{0009-0003-7928-3866}
\email{sakyachen@tencent.com}

\author{Bin Hu}
\orcid{0009-0007-7143-1763}
\email{glennbinhu@tencent.com}
\affiliation{
\institution{Tencent}
\city{Shenzhen}
\country{China}
}

\author{Lili Huang}
\orcid{0009-0006-7282-7084}
\email{lilillhuang@tencent.com}

\author{Yanwen Wu}
\orcid{0009-0002-8129-9528}
\email{steveywwu@tencent.com}

\author{Yeshou Cai}
\orcid{0009-0009-9067-479X}
\email{showcai@tencent.com}
\affiliation{
\institution{Tencent}
\city{Shenzhen}
\country{China}
}

\author{Qi Zhou}
\orcid{0009-0005-3579-1177}
\email{dzhouqi@gmail.com}

\author{Huang Tang}
\orcid{0009-0001-1981-1722}
\email{michaeltang@tencent.com}

\author{Chunfeng Yang}
\orcid{0000-0002-0324-8541}
\email{yannisyang@tencent.com}
\affiliation{
\institution{Tencent}
\city{Shenzhen}
\country{China}
}

\author{Chengguo Yin}
\orcid{0009-0008-0779-9579}
\email{turingyin@tencent.com}

\author{Tingyu Jiang}
\orcid{0009-0000-7642-0755}
\email{travisjiang@tencent.com}

\author{Lifeng Wang}
\orcid{0009-0007-4353-0721}
\email{327939898@qq.com}
\affiliation{
\institution{Tencent}
\city{Shenzhen}
\country{China}
}

\author{Shudong Huang}
\orcid{0009-0007-6215-272X}
\authornotemark[2]
\email{ericdhuang@tencent.com}
\affiliation{
\institution{Tencent}
\city{Shenzhen}
\country{China}
}

\author{Dapeng Liu}
\orcid{0009-0003-2973-9167}
\email{rocliu@tencent.com}

\author{Yuekui Yang}
\orcid{0000-0002-8709-3128}
\email{yuekuiyang@tencent.com}
\affiliation{
\institution{Tencent}
\city{Shenzhen}
\country{China}
}

\author{Shu-Tao Xia}
\orcid{0000-0002-8639-982X}
\email{xiast@sz.tsinghua.edu.cn}
\affiliation{
\institution{Tsinghua University}
\city{Shenzhen}
\country{China}
}

\author{Lei Xiao}
\orcid{0009-0002-3991-8161}
\email{shawnxiao@tencent.com}

\author{Haijie Gu}
\orcid{0009-0001-4424-2322}
\email{jerrickgu@tencent.com}
\affiliation{
\institution{Tencent}
\city{Shenzhen}
\country{China}
}

\renewcommand{\shortauthors}{Jie Jiang et al.}

%% file: 1.introduction.tex
\section{Introduction}

\begin{figure}
\includegraphics[width=\linewidth]{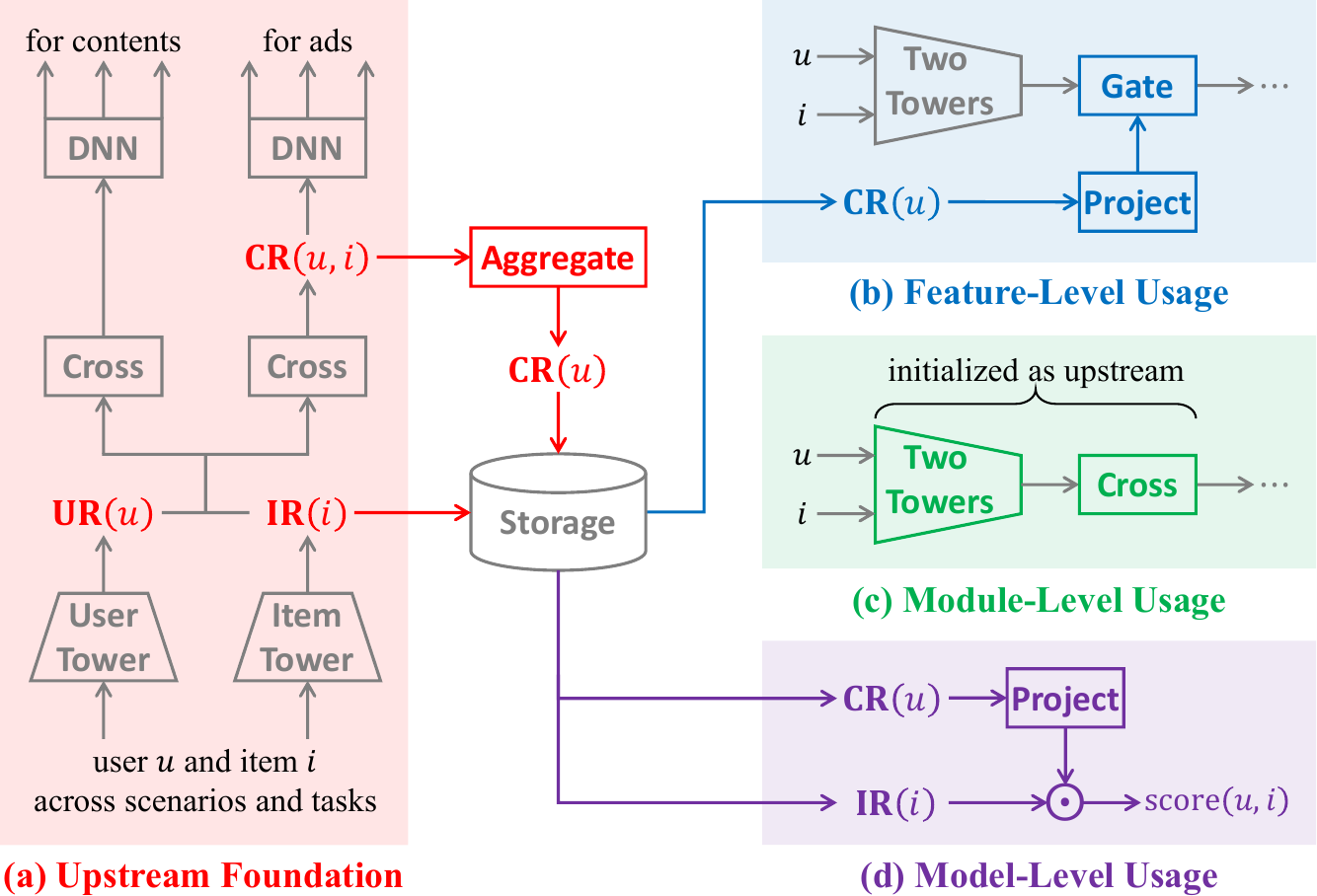}
\caption{
Overview.
(a) LFM4Ads is pretrained on content and ad samples across scenarios and tasks, and representations are transferred to downstream models.
(b) Cross representations are fused by gating for feature-level usage.
(c) Modules that yield cross representations are fine-tuned for module-level usage.
(d) Cross and item representations are matched for model-level usage.
}
\label{fig:lfm4ads-overview}
\end{figure}

Online advertising has grown into a massive digital economy, driving hundreds of billions of dollars in annual revenue~\cite{iab2025internet}. Its ranking and bidding systems depend on click-through rate (CTR) and conversion rate (CVR) prediction, motivating extensive research on multi-field feature interactions~\cite{rendle2010factorization,cheng2016wide,pan2018fieldweighted} and sequential behavior modeling~\cite{zhou2019deep,zhou2024temporal,feng2025longsequence}. As platforms span many scenarios and tasks, multi-task and multi-domain learning can share statistical strength~\cite{pan2019predicting,su2024stem,lin2025crocodile}, but may suffer from negative transfer and the seesaw effect~\cite{ma2018modeling,tang2020progressive,zhou2023hinet,bai2022contrastive}. Strict latency, throughput, and cost constraints also make a single large joint model difficult to deploy. This motivates the \emph{pretrain-then-transfer} paradigm, where foundation models (FMs) learn reusable knowledge from cross-scenario data and adapt it to specialized downstream models~\cite{li2026realizing,liang2025external,zhou2026generative,pancha2022pinnerformer,wang2024empowering,zhang2024scaling,chen2025pinfm,li2026fmpd}.

However, existing transfer pipelines remain incomplete in both \emph{what} is transferred and \emph{how} it is utilized.
First, most methods transfer user representations (URs) and item representations (IRs), typically obtained from upstream user or item towers~\cite{pancha2022pinnerformer,zhang2024scaling,pan2024ads}.
Although URs and IRs can capture user-side and item-side characteristics, they underutilize finer-grained user-item correlations generated by deeper interaction layers.
A recent work transfers Target-Aware Embeddings (TAEs) of user-item pairs~\cite{li2026realizing}, but it requires the FM to compute TAEs in real time, resulting in an additional upstream inference alongside the downstream.
Efficient transfer of user-item cross representations (CRs) therefore remains unresolved.

Second, the transferability of CRs is underexplored. Modern recommenders contain multiple explicit and implicit interaction layers~\cite{sun2021fm2,wang2021dcn,kang2025unifying}, but it is unclear which layers preserve the most generalizable CRs.
Moreover, CRs are sample-specific and lack reusable coarse-grained forms for downstream tasks. Additionally, existing transfer mechanisms are largely confined to feature-level injection through simple projections, which may fail to bridge upstream-downstream semantic gaps and may dilute upstream signals among thousands of downstream features.
Module-level reuse of pretrained modules and model-level retrieval using pretrained representations remain underexplored.

To bridge these gaps, we propose Large Foundation Model for Ads Recommendation (LFM4Ads), a novel \emph{pretrain-then-transfer} paradigm for cross-domain ad recommendation, as illustrated in Figure~\ref{fig:lfm4ads-overview}. Our contributions include:
\begin{itemize}
\item We extract all available representations from the FM and leverage their complementary strengths: URs record user interests, IRs encode item-side information, and CRs provide finer-grained user-item interaction signals.
\item We investigate CRs from diverse interaction layers and aggregate sample-level CRs into user-level CRs using a time-aware exponential moving average, making CRs reusable for downstream tasks.
\item We transfer upstream knowledge at the feature, module, and model levels. Specifically, we fuse CRs via a non-linear gate, reuse upstream modules that yield CRs, and match CRs with IRs. From feature-level to module-level and then to model-level, our transfer granularity shifts from fine to coarse, and the downstream training burden shifts from heavy to light, demonstrating that \emph{the stronger the foundation model, the thinner the downstream models.}
\end{itemize}

We comprehensively evaluate LFM4Ads across eight scenarios in a public dataset, KuaiRand-1K.
Our feature-level gating, module-level fine-tuning, and model-level retrieval improve the average AUC by 0.0025, 0.0016, and 0.0189 over baselines, respectively.
In online deployment,
LFM4Ads has supported 10+ production launches and yields a platform-wide gross merchandise value (GMV) lift of 2.45\%, translating to annual revenue increases in the hundreds of millions of dollars.
Specifically, feature-level fusion increases GMV by 1.35\% in Weixin Moments,
module-level fine-tuning by 1.88\% across multiple pCVR models, and
model-level retrieval by 1.17\% in Weixin Channels.

%% file: 2.background.tex
\section{Related Works}

Industrial advertising and recommendation models learn user-item relevance using complementary techniques, including multi-field feature interaction modeling~\cite{cheng2016wide,guo2017deepfm,wang2021dcn}, multi-task learning~\cite{pan2019predicting}, sequential behavior modeling~\cite{zhou2019deep,zhou2024temporal,feng2025longsequence}, and vertical federated learning~\cite{li2024refer,li2026prefer,li2026jpl}. Together, these techniques underpin modern CTR and CVR prediction pipelines. Although these models capture high-order interactions and evolving user interests, strict constraints on latency, computational resources, and feature budgets limit their scale in online serving~\cite{grbovic2018realtime,pi2019practice,guo2024embedding}.

To address these constraints, upstream-downstream systems pretrain large models on diverse data and transfer knowledge through representation injection or model adaptation. Representative systems learn sequence-based user embeddings~\cite{pancha2022pinnerformer,zhang2024scaling}, transfer multi-modal item signals~\cite{baltescu2022itemsage,sheng2024enhancing}, or reuse learned user-item similarities~\cite{pan2024ads}. Model-adaptation methods fine-tune models pretrained on user activity sequences~\cite{chen2025pinfm} or distill knowledge from an external foundation model into smaller domain-specific models~\cite{liang2025external}. Despite these advances, sample-specific interaction knowledge from deeper layers remains difficult to store, transfer, and reuse across heterogeneous downstream architectures.

A parallel line of research explores large language model (LLM)-based approaches to e-commerce instruction tuning~\cite{li2024ecomgpt,peng2024ecellm,shi2025llamae} and generative, cross-domain, or generalizable recommendation~\cite{wang2024empowering,xin2025llmcdsr,zhou2026generative}. Recent work also studies dimensional degeneration in generative ranking models with LLM-inspired architectures~\cite{xin2026collapse}.

%% file: 3.method.tex
\section{Methodology}

We first design our upstream foundation model in Section~\ref{sec:pretrain-representations}, then enhance cross representations in Section~\ref{sec:cr-transferability}, and finally discuss three downstream usages in Section~\ref{sec:downstream-usages}.

\subsection{Pretrain and Extract Representations}
\label{sec:pretrain-representations}

Our Large Foundation Model for Ads Recommendation (LFM4Ads) is illustrated in Figure~\ref{fig:lfm4ads-overview}(a).
We adopt a \emph{two-tower} design, where a user tower extracts a user representation (UR) from user features, and an item tower extracts an item representation (IR) from item features.
Then we employ interaction layers (e.g., DCNv2~\cite{wang2021dcn}) to capture the user-item cross correlations.

Formally, let $\textbf{UR}(u)$ be the representation of user $u$, and $\textbf{IR}(i)$ be the representation of item $i$.
Whenever a sample $(u,i)$ arrives at LFM4Ads, we extract $\textbf{UR}(u)$ and $\textbf{IR}(i)$ from the two towers.
Besides, the interaction layers capture the correlations between $\textbf{UR}(u)$ and $\textbf{IR}(i)$ at finer-grained levels, including explicit and implicit feature interactions.
To leverage these interactions, we extract the output of an interaction layer as a user-item cross representation (CR), denoted as $\textbf{CR}(u,i)$.
\emph{Together, URs, IRs, and CRs constitute the three representation types transferred in LFM4Ads}.

LFM4Ads is pretrained on all samples from multiple scenarios (e.g., Weixin Moments, Search Ads), multiple tasks (e.g., pCTR, pCVR), and two domains (content and advertising).
To account for distributional differences between content and ad samples, we adopt a \emph{dual-branch} design after the two towers, with one branch for content tasks and the other for ad tasks.
Notably, we transfer CRs only from the ad branch because our downstream tasks are ad recommendations.
For content tasks, we suggest transferring the CRs from the content branch.

\subsection{Enhance Transferability of CRs}
\label{sec:cr-transferability}

There are two main challenges in transferring CRs:
(1) CR is a sample-level representation of a user-item pair, which is too fine-grained for downstream tasks.
(2) It is impractical to calculate and store the CRs for all user-item pairs.
Although a recent work~\cite{li2026realizing} computes Target-Aware Embeddings (TAEs) in real time, it requires an additional upstream inference alongside the downstream.

To address these challenges, we aggregate sample-level CRs for each user into one user-level CR.
Formally, let $\textbf{CR}(u)$ be the aggregated CR for user $u$.
We initialize $\textbf{CR}(u)\gets\boldsymbol{0}$.
Whenever a new $\textbf{CR}(u,i)$ is generated, we update $\textbf{CR}(u)$ by
\begin{align}
\label{eq:cr-aggregation}
\textbf{CR}(u)\gets\beta(t)\cdot\textbf{CR}(u)+\big(1-\beta(t)\big)\cdot\textbf{CR}(u,i),
\end{align}
where $t$ is the time interval since the previous update of $\textbf{CR}(u)$, and $\beta:\mathbb{R}_+\to[0,1]$ is a decreasing function.

We refer to (\ref{eq:cr-aggregation}) as a \emph{time-aware exponential moving average}, which adapts to user activity.
For inactive users, $t$ is large and we have $\beta(t)\approx 0$ to prioritize current interactions and address distribution drift.
For active users, $t$ is small and we have $\beta(t)\approx 1$ to retain more historical interactions.

To further enhance the transferability of CRs, we explore which layer yields the best CRs.
We find that the optimal CRs are extracted from the middle layer for feature-level usage, and from the deepest layer for model-level usage.
More details are in Sections~\ref{sec:offline-feature}~and~\ref{sec:offline-model}.

\subsection{Three Downstream Usages}
\label{sec:downstream-usages}

In this section, we discuss how to use CRs for downstream tasks.
We propose feature-level, module-level, and model-level usages, ranging from fine-grained to coarse-grained.

\subsubsection{Feature-Level Usage: Non-linear Gating}

A straightforward fusion method is to treat an upstream representation as an additional feature and concatenate it with downstream embeddings, like SUM~\cite{zhang2024scaling}. Formally, let $\boldsymbol{E}$ be the concatenation of downstream embeddings.
The concat-based fusion yields
\begin{align}
\label{eq:concat-fusion}
\boldsymbol{E}'=\big[\boldsymbol{E},\quad\boldsymbol{M}\cdot\textbf{CR}(u)\big],
\end{align}
where $\boldsymbol{M}$ is a learnable projection matrix that reduces the length of $\textbf{CR}(u)$.

However, (\ref{eq:concat-fusion}) may be ineffective because it directly combines $\boldsymbol{E}$ and $\textbf{CR}(u)$ which come from two semantically misaligned spaces.
We tried several approaches, including employing an alignment loss between the upstream and downstream features~\cite{li2025ctrl}, but none of them led to a performance gain over production baselines during online A/B tests.
Inspired by self-gating~\cite{hu2018squeezeandexcitation,chang2023pepnet,yin2025feature}, we adopt a gating-based fusion, which yields
\begin{align}
\label{eq:gated-fusion}
\boldsymbol{E}'=\boldsymbol{E}\odot\boldsymbol{\sigma}\big(\boldsymbol{M}\cdot\textbf{CR}(u)\big),
\end{align}
where $\boldsymbol{\sigma}$ is a non-linear function and $\odot$ is the Hadamard product.
$\boldsymbol{M}$ projects $\textbf{CR}(u)$ to the same length as $\boldsymbol{E}$, without increasing the dimensionality of subsequent layers.
Here $\boldsymbol{\sigma}(\boldsymbol{M}\cdot)$ serves as a gate as illustrated in Figure~\ref{fig:lfm4ads-overview}(b).
A recent work~\cite{yin2025feature} revealed the mechanism behind (\ref{eq:gated-fusion}) ---
$\boldsymbol{\sigma}(\boldsymbol{M}\cdot)$ \emph{serves as a non-linear encoder to construct new representations, thereby mitigating dimensional collapse and representation redundancy.}

\subsubsection{Module-Level Usage: Fine-Tuning Cross Modules}

As discussed in Section~\ref{sec:cr-transferability}, there are challenges in transferring upstream $\textbf{CR}(u,i)$.
To overcome them, we have proposed transferring the aggregated $\textbf{CR}(u)$.
In this section, we propose another solution --- transferring the upstream modules that yield $\textbf{CR}(u,i)$.

As illustrated in Figure~\ref{fig:lfm4ads-overview}(c), we initialize the downstream model with the same modules as LFM4Ads, including the two towers and the interaction layers.
Thus, the downstream model yields the same $\textbf{CR}(u,i)$ as LFM4Ads.
We then fine-tune these modules on downstream samples to yield $\textbf{CR}'(u,i)$, a fine-tuned cross representation better aligned with downstream objectives.

Prior works have shown that pretrained representations can transfer user-item similarity structures across models~\cite{sheng2024enhancing,pan2024ads,pan2025computational}.
Building on these findings, we transfer the upstream modules so that downstream models inherit the user-item correlations learned by LFM4Ads.
This makes downstream learning start from a representation space where user-item interactions have already been shaped by multi-scenario pretraining.

\subsubsection{Model-Level Usage: Standalone Retrieval}
URs and IRs learned by LFM4Ads naturally encode the relevance between users and items, so they can be directly used for retrieval~\cite{pancha2022pinnerformer}.
Formally, given a user $u$, we retrieve the best-matching item $i$ from the candidate pool, with the highest
\begin{align}
\label{eq:retrieval-score}
\text{score}(u,i)=\left<\textbf{Adapter}\big(\textbf{UR}(u)\big),\quad\textbf{IR}(i)\right>,
\end{align}
where $\left<\cdot,\cdot\right>$ is the dot product, and $\textbf{Adapter}$ is a tiny learnable layer (e.g., a projection matrix).
We optimize $\textbf{Adapter}$ using binary cross-entropy (BCE) or InfoNCE~\cite{oord2018representation}.

Since our $\textbf{CR}(u)$ can be regarded as a user representation like $\textbf{UR}(u)$, we explore replacing $\textbf{UR}(u)$ with $\textbf{CR}(u)$ in (\ref{eq:retrieval-score}), and find that $\textbf{CR}(u)$ generally performs better.
More details are in Section~\ref{sec:offline-model}.

As illustrated in Figure~\ref{fig:lfm4ads-overview}(d), our retrieval model (\ref{eq:retrieval-score}) does not require two parameter-heavy towers.
Benefiting from our effective and refined $\textbf{CR}(u)$, a tiny adapter is sufficient to serve as a standalone retrieval model.
This highlights the potential of LFM4Ads
---
\emph{it provides not only an additional representation, but also a standalone model.}

\subsubsection{Trade-off among Three Usages}

Feature-level usage is compatible with existing online ranking models, since it only injects an additional representation as a new feature.
It can provide performance gains for well-trained mature systems.

Module-level usage is suitable for new ranking models.
It not only accelerates training convergence but also leverages full-scenario knowledge to tackle long-tail scenarios.
But it lacks flexibility, since it requires upstream and downstream models to share similar architectures.

Model-level usage is convenient for fast deployment.
With few parameters and low latency, it is suitable for large-scale retrieval.
However, model-level usage achieves lower AUC than the other two usages, due to fewer input features and smaller model capacity.

From feature-level to module-level and then to model-level, \emph{the transfer granularity ranges from fine to coarse, and thus the downstream training burden shifts from heavy to light.}

%% file: 4.offline.tex
\section{Offline Experiments}

There is no widely adopted benchmark for evaluating recommendation foundation models.
We therefore use KuaiRand-1K~\cite{gao2022kuairand}, which contains interactions spanning 15 scenarios, to emulate a \emph{pretrain-then-transfer} paradigm. Specifically, we pretrain an upstream foundation model on the union of data from all scenarios, and then transfer it to scenario-specific downstream models, each trained on samples from only its own scenario.

\subsection{Settings}

\subsubsection{Dataset}
KuaiRand-1K contains 1,000 users, 4,369,953 items, and 11,713,045 samples.
These samples cover 15 scenarios and 31 days.
We use samples from the first 25 days for training, the following 3 days for validation, and the last 3 days for testing.
Each downstream dataset contains samples from only one scenario.
We discard continuous features and retain 27 user-side and 9 item-side categorical features for each sample.
We take click or not as the label, constructing a pCTR task.

\subsubsection{Upstream Model}
Our foundation model comprises a user tower, an item tower, and a DCNv2~\cite{wang2021dcn}. The DCNv2 has 3 cross layers and a 3-layer DNN.
All layers are sequentially denoted as
\begin{align*}
Towers,\ Cross_1,\ Cross_2,\ Cross_3,\ DNN_1,\ DNN_2,\ DNN_3.
\end{align*}
$Towers$ map each feature to an embedding of length 10, and then concatenate them.
The size of $Cross_1$, $\cdots$, $DNN_2$ is $360\times360$, while the size of $DNN_3$ is $360\times15$.
The foundation model consumes samples from all 15 scenarios, emulating online cross-scenario pretraining.
The output is 15 logits for 15 scenarios, but only the logit for the sample's scenario is used.

\subsubsection{Training}

Our foundation and downstream models share the same training settings.
The loss is BCE and the optimizer is AdamW~\cite{loshchilov2019decoupled}.
The settings of AdamW follow PyTorch defaults, where the learning rate is 0.001, betas are (0.9, 0.999), and weight decay is 0.01.
The samples are sorted by timestamp and the batch size is 10,000.
We stop training when the validation AUC stops increasing.

\subsubsection{Upstream Representations}

After pretraining, we freeze the foundation model and transfer $\textbf{UR}(u)$ and $\textbf{CR}(u)$ for each user $u$, and $\textbf{IR}(i)$ for each item $i$.
$\textbf{UR}(u)$ and $\textbf{IR}(i)$ are extracted from $Towers$.
To obtain $\textbf{CR}(u)$, we replay the training and validation samples.
For each sample $(u,i)$, we extract $\textbf{CR}(u,i)$ from $Towers$, $Cross_1$, $\cdots$, or $DNN_2$,
then aggregate sample-level $\textbf{CR}(u,i)$ into user-level $\textbf{CR}(u)$ using (\ref{eq:cr-aggregation}) with $\beta=0.99$.

\subsubsection{Evaluation} We evaluate the test AUC of each downstream model, which is trained on samples from only a specific scenario.
We average the test AUC over 100 runs, each involving an individual pair of foundation and downstream models.

\subsection{Feature-Level Usage}
\label{sec:offline-feature}

\begin{figure*}
\begin{subfigure}[t]{0.32\textwidth}
\includegraphics[height=4cm]{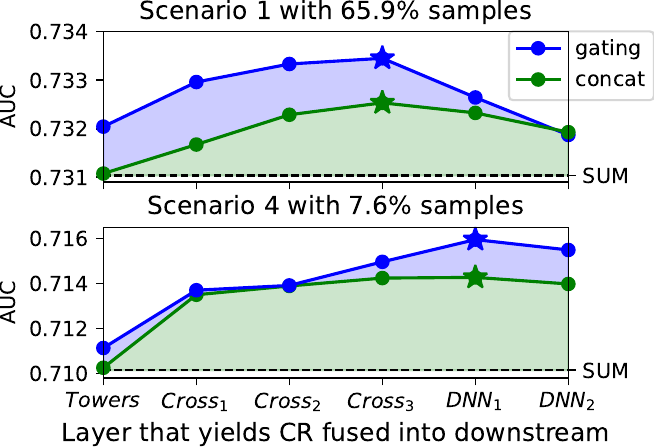}
\caption{
Feature-level usage.
SUM fuses UR via concat~\cite{zhang2024scaling}.
{\color[RGB]{204,230,204}\rule{5pt}{5pt}} indicates the gain from CR, and {\color[RGB]{204,204,255}\rule{5pt}{5pt}} indicates the gain from gating.
}
\end{subfigure}
\hfill
\begin{subfigure}[t]{0.32\textwidth}
\includegraphics[height=4cm]{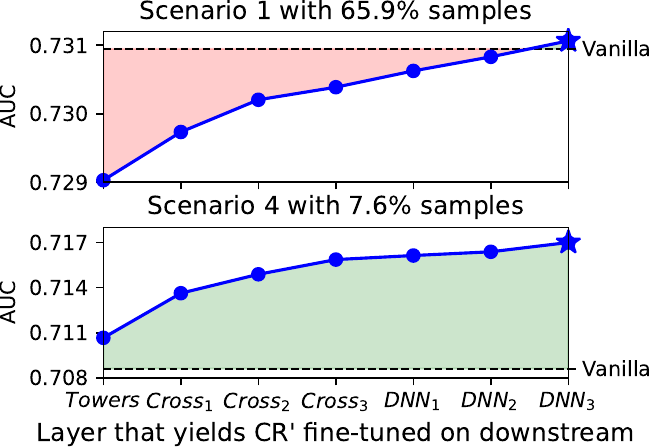}
\caption{
Module-level usage.
Vanilla trains from scratch.
{\color[RGB]{204,230,204}\rule{5pt}{5pt}} indicates the gain from upstream initialization, while
{\color[RGB]{255,204,204}\rule{5pt}{5pt}} indicates the drop.
}
\end{subfigure}
\hfill
\begin{subfigure}[t]{0.32\textwidth}
\includegraphics[height=4cm]{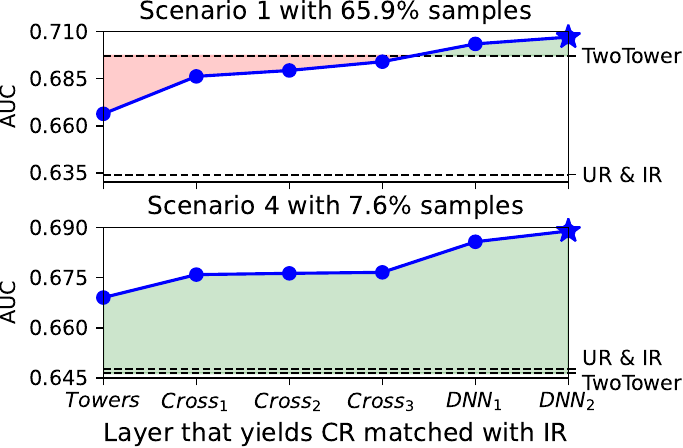}
\caption{
Model-level usage.
TwoTower uses downstream representations.
{\color[RGB]{204,230,204}\rule{5pt}{5pt}} indicates the gain from CR \& IR, while {\color[RGB]{255,204,204}\rule{5pt}{5pt}} indicates the drop.
}
\end{subfigure}
\caption{Offline results on two typical scenarios of KuaiRand-1K.}
\label{fig:offline-transfer-results}
\end{figure*}

We train downstream models with the same architecture as the upstream FM and fuse $\textbf{CR}(u)$ into the downstream embeddings.
For concat-based fusion (\ref{eq:concat-fusion}), the length of $\textbf{CR}(u)$ is reduced to 10 to align with a downstream embedding.
For gating-based fusion (\ref{eq:gated-fusion}), we let $\boldsymbol{\sigma}=\textbf{sigmoid}$ to introduce non-linearity.
As a baseline, SUM fuses $\textbf{UR}(u)$ via concat~\cite{zhang2024scaling}.

Figure~\ref{fig:offline-transfer-results}(a) compares feature-level usages in Scenarios~1~and~4, which are the largest and a typical medium scenario, respectively.
Concatenating CRs generally outperforms SUM as indicated by {\color[RGB]{204,230,204}\rule{5pt}{5pt}}, and gating with CRs further improves AUC as indicated by {\color[RGB]{204,204,255}\rule{5pt}{5pt}}.
With CRs extracted from $Cross_3$, concat (\ref{eq:concat-fusion}) improves AUC over SUM by 0.0015 and 0.0041 in Scenarios~1~and~4, respectively, and
gating (\ref{eq:gated-fusion}) yields further gains of 0.0009 and 0.0007 over concat.
Note that a 0.001 AUC gain can be practically meaningful in the recommendation community~\cite{li2024simcen,liu2024multifs}.
These gains indicate that CRs provide useful information about finer-grained user-item interactions and gating helps downstream models leverage this information.
Notably, concat (\ref{eq:concat-fusion}) increases the dimensionality of subsequent layers, whereas gating (\ref{eq:gated-fusion}) does not.

In both scenarios, AUC first increases and then decreases as the layer for extracting CRs becomes deeper.
The highest AUC is achieved at $Cross_3$ in Scenario~1 and $DNN_1$ in Scenario~4, both of which are intermediate layers.
This pattern indicates a balance between representation refinement and task specificity, consistent with~\cite{yosinski2014how}:
Shallow CRs may be insufficiently processed for downstream use, whereas deeper CRs may be too specialized to upstream tasks.
Intermediate CRs provide useful interaction information while retaining flexibility for downstream tasks.

\subsection{Module-Level Usage}

We train downstream models with the same architecture as the upstream FM.
The preceding layers are initialized with upstream weights, while the remaining layers are initialized from scratch.
As a baseline, Vanilla initializes all layers from scratch.

Figure~\ref{fig:offline-transfer-results}(b) shows that AUC increases monotonically as more layers are transferred, with the highest AUC achieved when all layers are transferred in both scenarios.
This trend indicates that transferring more layers provides more useful interaction knowledge learned by the FM.

However, the gains over Vanilla also vary with scenario size.
In Scenario~1 with 65.9\% samples, partial transfer leads to negative effects as indicated by {\color[RGB]{255,204,204}\rule{5pt}{5pt}}.
In Scenario~4 with 7.6\% samples, each transfer depth outperforms Vanilla as indicated by {\color[RGB]{204,230,204}\rule{5pt}{5pt}}.
Our explanation is that, when downstream samples are abundant, the model can learn effectively from scratch.
Partial transfer may introduce the bias from the FM, but lacks subsequent correction, resulting in negative effects.

\begin{table*}
\caption{
Offline results on all available scenarios of KuaiRand-1K.
Highest is bolded and lowest is italicized.
Average is weighted by the number of samples.
CR is extracted from $Cross_3$ for feature-level usage, and $DNN_2$ for model-level usage. All layers are fine-tuned to yield $\textbf{CR}'$ for module-level usage.
}
\begin{tabular}{ccccccccccc}
\toprule
\multicolumn{2}{c}{\textbf{Scenario}}
&\textbf{1}&\textbf{0}&\textbf{4}&\textbf{2}&\textbf{6}&\textbf{3}&\textbf{8}&\textbf{5}&\textbf{Average}\\
\midrule
\multicolumn{2}{c}{\textbf{\# of samples}}
&7,717,601&2,407,352&895,385&402,293&183,403&37,418&17,430&16,424&--\\
\multicolumn{2}{c}{\textbf{\% of samples}}
&65.9\%&20.6\%&7.6\%&3.4\%&1.6\%&0.3\%&0.1\%&0.1\%&--\\
\midrule
\multirow{3}{*}{Feature}
&\textbf{SUM}&\textit{0.7310}&\textit{0.7123}&\textit{0.7101}&\textit{0.7975}&\textit{0.7183}&\textit{0.7017}&\textit{0.7622}&\textit{0.6562}&\textit{0.7275}\\
&\textbf{concat CR}&0.7325&0.7123&0.7142&0.7993&\textbf{0.7277}&0.7114&\textbf{0.7668}&0.6622&0.7291\\
&\textbf{gating CR}&\textbf{0.7334}&\textbf{0.7132}&\textbf{0.7150}&\textbf{0.8000}&0.7249&\textbf{0.7302}&0.7657&\textbf{0.6699}&\textbf{0.7300}\\
\midrule
\multirow{2}{*}{Module}
&\textbf{Vanilla}&\textit{0.7309}&\textit{0.7120}&\textit{0.7086}&\textit{0.7971}&\textit{0.7173}&\textit{0.7038}&\textit{0.7627}&\textit{0.6606}&\textit{0.7273}\\
&$\textbf{CR}'$&\textbf{0.7311}&\textbf{0.7136}&\textbf{0.7170}&\textbf{0.7990}&\textbf{0.7361}&\textbf{0.7387}&\textbf{0.7861}&\textbf{0.6834}&\textbf{0.7288}\\
\midrule
\multirow{3}{*}{Model}
&\textbf{TwoTower}&0.6972&0.6450&\textit{0.6464}&\textit{0.7570}&\textit{0.5776}&\textit{0.5838}&\textbf{0.6550}&\textit{0.5562}&0.6821\\
&\textbf{UR \& IR}&\textit{0.6340}&\textit{0.6376}&0.6478&\textbf{0.7662}&0.5915&0.5911&0.6449&0.5565&\textit{0.6395}\\
&\textbf{CR \& IR}&\textbf{0.7070}&\textbf{0.6821}&\textbf{0.6889}&0.7645&\textbf{0.6553}&\textbf{0.6122}&\textit{0.5646}&\textbf{0.6012}&\textbf{0.7010}\\
\bottomrule
\end{tabular}
\label{tab:offline-results}
\end{table*}

\subsection{Model-Level Usage}
\label{sec:offline-model}

We instantiate the $\textbf{Adapter}$ in (\ref{eq:retrieval-score}) as a learnable matrix $\boldsymbol{W}$:
\begin{align}
\label{eq:ur-ir-score}
\text{score}(u,i)=\textbf{UR}^T(u)\cdot\boldsymbol{W}\cdot\textbf{IR}(i).
\end{align}
We also try a variant that replaces URs with CRs:
\begin{align}
\label{eq:cr-ir-score}
\text{score}(u,i)=\textbf{CR}^T(u)\cdot\boldsymbol{W}\cdot\textbf{IR}(i).
\end{align}
As a baseline, we train a model with its own two towers:
\begin{align}
\label{eq:two-tower-score}
\text{score}(u,i)=\textbf{UserTower}^T(u)\cdot\boldsymbol{W}\cdot\textbf{ItemTower}(i),
\end{align}
where $\textbf{UserTower}$ and $\textbf{ItemTower}$ have the same architecture as FM's, and are trained from scratch on downstream samples.
We denote (\ref{eq:ur-ir-score}) as UR \& IR, (\ref{eq:cr-ir-score}) as CR \& IR, and (\ref{eq:two-tower-score}) as TwoTower, respectively.

Figure~\ref{fig:offline-transfer-results}(c) compares the three retrieval models.
TwoTower outperforms UR \& IR in Scenario~1, while UR \& IR slightly outperforms TwoTower in Scenario~4.
These results indicate that abundant downstream samples support learning effective two towers from scratch, whereas upstream representations perform better when downstream supervision is limited.
Additionally, with CRs extracted from $DNN_2$, CR \& IR outperforms both baselines as indicated by {\color[RGB]{204,230,204}\rule{5pt}{5pt}}.
Its AUC gains over TwoTower are 0.0099 and 0.0425 in Scenarios~1~and~4, respectively, and those over UR \& IR are 0.0730 and 0.0411.
These further gains indicate that CRs capture useful user-item interaction information for downstream retrieval.

In both scenarios, AUC increases monotonically as the layer for extracting CRs becomes deeper and peaks at $DNN_2$, the deepest layer.
This pattern may arise because CR \& IR trains only a matrix adapter and thus relies on the more refined knowledge provided by deeper upstream layers.
Notably, CRs and IRs allow us to avoid training the two towers, substantially reducing the number of downstream parameters.

\subsection{Overall Comparisons}

We conduct comprehensive experiments on the eight scenarios with sufficient test samples in KuaiRand-1K.
Following the best practices in Sections~\ref{sec:offline-feature}--\ref{sec:offline-model}, we extract CRs from $Cross_3$ for feature-level usage, initialize all layers with upstream weights for module-level usage, and extract CRs from $DNN_2$ for model-level usage.

The results listed in Table~\ref{tab:offline-results} show that all three usages generally improve AUC over their respective baselines.
For the average AUC weighted by the number of samples, feature-level gating yields a gain of 0.0025 over SUM,
module-level fine-tuning yields 0.0016 over Vanilla,
and model-level retrieval yields 0.0189 over TwoTower.

Notably, the AUC gains are particularly substantial in Scenarios~3,~8,~and~5, which together account for only 0.6\% samples.
Specifically,
feature-level gating improves AUC by 0.0284 in Scenario~3,
module-level fine-tuning by 0.0233 in Scenario~8, and
model-level retrieval by 0.0450 in Scenario~5.
These results indicate that pretraining on full-scenario data provides important knowledge that is difficult to learn from limited downstream samples, making transfer especially valuable for long-tail scenarios.

%% file: 5.online.tex
\section{Online Deployment}

We instantiate LFM4Ads as an industrial system, and deploy it on Tencent's advertising platform at scale.

\subsection{Samples and Features}

We collect samples from all available advertising scenarios, including Weixin Moments, Weixin Channels, Weixin Official Accounts, Tencent News, Tencent Sports, Demand-Side Platform, and other pCTR and pCVR scenarios.
Besides advertisements, we also collect samples from the content domain, which is on a much larger scale.
We trained a dedicated multi-modal LLM to annotate content samples with commercial labels, aligning them with ad samples.
The total number of samples fed into LFM4Ads is tens of billions per day, with 80\% from the content domain and 20\% from the ad domain.
The number of samples is 20 to hundreds of times larger than that of individual downstream tasks.

Each upstream sample contains approximately 1,800 features, most of which are user-side.
In contrast, most downstream ranking samples have 200--1,000 features, while matching and retrieval samples have 100--500 features.

87\% of the upstream user features are multi-valued, including around 50 sequences, with lengths ranging from dozens to thousands.
On average, flattening all multi-valued features yields more than 60,000 values per sample.

\subsection{Scale of Models}

LFM4Ads has a 4-TB parameter footprint, which is 48\% larger than the Weixin Moments pCTR model, 233\% larger than the Weixin Channels pCTR model, and 427\% larger than the average size of downstream models.
It processes 6.3B sparse features, 270\% more than the largest downstream task processes.
The FLOPS is 1.45B and the QPS is 500K.
To save storage, we employ quantization to compress an upstream representation from 4KB to 1KB.

\subsection{Response Process}

\begin{figure}
\includegraphics[width=\linewidth]{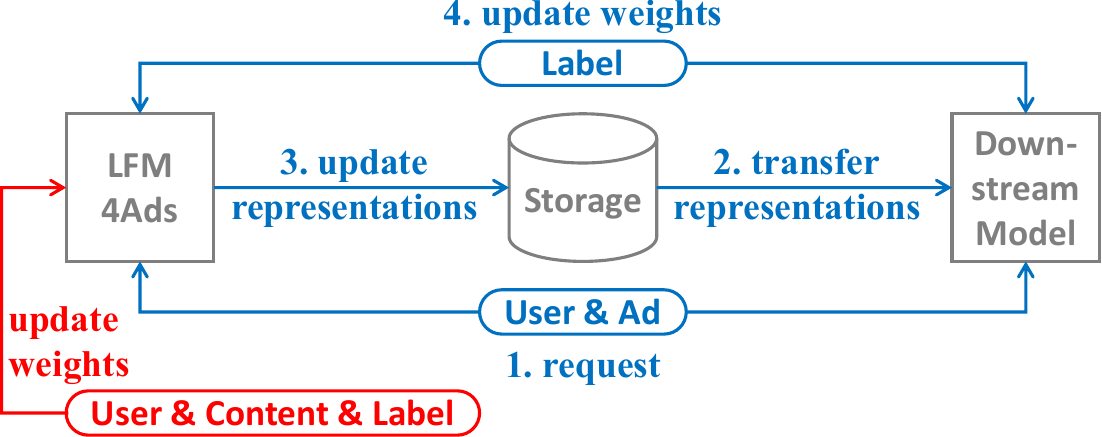}
\caption{Online response process.}
\label{fig:online-response}
\end{figure}

The response process of our online system is illustrated in Figure~\ref{fig:online-response}.
Formally, whenever a user $u$ and an ad $i$ arrive at the downstream, we retrieve $\textbf{UR}(u)$, $\textbf{IR}(i)$, and $\textbf{CR}(u)$ from the storage, and transfer them into the downstream model. Meanwhile, LFM4Ads infers on $(u,i)$ and yields new $\textbf{UR}(u)$, $\textbf{IR}(i)$ and $\textbf{CR}(u,i)$. Then in the storage, $\textbf{UR}(u)$ and $\textbf{IR}(i)$ are updated directly, and $\textbf{CR}(u)$ is aggregated by (\ref{eq:cr-aggregation}). Whenever the label arrives, i.e., the user reacts to the ad, we update the weights of LFM4Ads and the downstream model.
Additionally, LFM4Ads is updated irregularly by content samples.
Content samples are much denser than ad samples, enabling LFM4Ads to capture users' latest interests.

\subsection{Launches and Productization}

LFM4Ads has been productized in 10+ major online launches, lifting overall GMV by 2.45\% across the platform.
At the scenario level, Weixin Moments achieves a 2.16\% GMV lift and Weixin Channels a 1.18\% lift.
At the task level, pCTR models yield a 0.18\% GMV lift, pCVR models a 1.33\% lift, and pLTV models a 0.44\% lift.

\begin{table}
\caption{Online A/B test results.}
\begin{tabular}{cccc}
\toprule
\textbf{Usage}&\textbf{Scenario}&\textbf{Task}&\textbf{GMV}\\
\midrule
\multirow{6}{*}{Feature}
&Weixin Moments&pCTR&+0.42\%\\
&Weixin Moments&pCVR&+2.53\%\\
&Weixin Channels&matching&+0.70\%\\
&Search Ads&matching&+1.75\%\\
&Demand-Side Platform&pCTR&+0.76\%\\
&Internet Service&pLTV&+0.93\%\\
\midrule
\multirow{5}{*}{Module}
&Weixin Moments&pCVR&+0.92\%\\
&Weixin Channels&pCVR&+0.92\%\\
&Weixin Official Accounts&pCVR&+0.94\%\\
&Content Platform&pCVR&+1.59\%\\
&Audience Network&pCVR&+0.95\%\\
\midrule
Model
&Weixin Channels&retrieval&+1.17\%\\
\bottomrule
\end{tabular}
\label{tab:online-ab-tests}
\end{table}

Table~\ref{tab:online-ab-tests} lists the online A/B test results of the three usages.
Feature-level usage generally improves performance across diverse scenarios and tasks, with the largest GMV gain of +2.53\% on the Weixin Moments pCVR model, indicating substantial business impact.
For module-level usage, LFM4Ads modules have been productized in multiple pCVR models, contributing to an overall GMV lift of 1.88\%.
Additionally, it speeds up downstream convergence and reduces training costs.
For model-level usage, a standalone retrieval model achieves a +1.17\% GMV lift in Weixin Channels, while also increasing CTR by 1.83\%, CTCVR by 3.34\%, and average dwell time by 1.66\%, and reducing the quick-slip rate by 0.36\%.

\subsection{Online Ablation}

Table~\ref{tab:online-transfer-ablation} reports online ablations with Weixin Moments as the downstream.
Compared with the baseline, single-branch SUM lowers AUC by 0.0014, whereas dual-branch SUM improves AUC by 0.0005 and yields a 0.79\% GMV lift.
It indicates that separate branches help accommodate differences between content and ad domains.
LFM4Ads further improves AUC by 0.0005 over dual-branch SUM and achieves a 1.35\% GMV lift over the baseline, indicating the benefit of CRs beyond URs.

Removing historical aggregation by setting $\beta=0$ in (\ref{eq:cr-aggregation}) reduces AUC by 0.0012 relative to standard LFM4Ads.
Note that $\beta=0$ implies $\textbf{CR}(u)=\textbf{CR}(u,i)$ where $i$ is the previous item.
Removing non-linearity by setting $\boldsymbol{\sigma}$ to the identity function reduces AUC by 0.0004.
Both variants yield negative GMV lifts, although the latter improves AUC by 0.0006 over the baseline.
These results highlight the necessity of historical aggregation and non-linear gating.

\subsection{Cross-Domain Analysis}

We examine how upstream user behavior sequences from the content and ad domains contribute to downstream ad recommendation.
The upstream baseline relies on only non-sequential features that describe users' static attributes.
The upstream representations are transferred to the Weixin Channels pCTR model.

\begin{table}
\caption{
Online ablations on transfer methods.
}
\begin{tabular}{lcccc}
\toprule
\textbf{Transferred}&\textbf{Branch}&\textbf{Fusion}&\textbf{AUC}&\textbf{GMV}\\
\midrule
None&--&--&0.8515&--\\
\midrule
$\textbf{UR}(u)$&single&concat&0.8501&negative\\
$\textbf{UR}(u)$&dual&concat&0.8520&+0.79\%\\
\midrule
$\textbf{CR}(u),\beta=0$&dual&gating&0.8513&negative\\
$\textbf{CR}(u),\boldsymbol{\sigma}=\textbf{id}$&dual&gating&0.8521&negative\\
$\textbf{CR}(u)$&dual&gating&\textbf{0.8525}&\textbf{+1.35\%}\\
\bottomrule
\end{tabular}
\label{tab:online-transfer-ablation}
\end{table}

\begin{table}
\caption{Online ablations on upstream domains.}
\begin{tabular}{ccccc}
\toprule
&\shortstack{\textbf{Static}\\\textbf{Only}}&\shortstack{\textbf{+ Content}\\\textbf{Sequence}}&\shortstack{\textbf{+ Ad}\\\textbf{Sequence}}&\shortstack{\textbf{+ Both}\\\textbf{Sequences}}\\
\midrule
\textbf{AUC}&--&+0.031&+0.042&+0.052\\
\bottomrule
\end{tabular}
\label{tab:upstream-domain-ablation}
\end{table}

As listed in Table~\ref{tab:upstream-domain-ablation}, incorporating sequences from both the content and ad domains yields the largest AUC gain of 0.052.
This exceeds the gain from ad sequences alone by 0.010 and that from content sequences alone by 0.021.
These results highlight that content behavior provides complementary information for ad prediction when combined with ad behavior.

%% file: 6.conclusion.tex
\section{Conclusion}

LFM4Ads is a novel pretrain-then-transfer paradigm for cross-domain ad recommendation, transferring upstream user, item, and cross representations at multiple granularities.
Feature-level gating, module-level fine-tuning, and model-level retrieval accommodate different downstream architectures and deployment requirements.
Offline experiments on KuaiRand-1K and production deployment on Tencent's advertising platform validate the effectiveness of LFM4Ads.
Extending transferable recommendation representations to downstream multi-channel bidding with explicit budget and ROI constraints remains an important direction~\cite{wu2026grb,guo2026l2fob}.

\begin{acks}
This work is supported in part by the \grantsponsor{GS501100001809}{National Natural Science Foundation of China}{https://doi.org/10.13039/501100001809} under Grant \grantnum{GS501100001809}{624B2088}.
\end{acks}

\section*{GenAI Usage Disclosure}

AI assisted only with manuscript language, grammar, and formatting.
All technical content, code, results, claims, and conclusions were independently developed by the authors.